\documentclass[lettersize,journal]{IEEEtran}
\usepackage{amsmath,amsfonts}
\usepackage{array}
\usepackage[caption=false,font=normalsize,labelfont=sf,textfont=sf]{subfig}
\usepackage{textcomp}
\usepackage{stfloats}
\usepackage{url}
\usepackage{verbatim}
\usepackage{graphicx}
\usepackage{cite}
\usepackage{graphicx}
\usepackage{subfig}
\usepackage{booktabs}
\usepackage{float}

\usepackage[utf8]{inputenc}
\usepackage{algpseudocode}
\usepackage{amsmath}

\usepackage{orcidlink}
\usepackage{multirow}
\usepackage{caption}
\usepackage{color, colortbl}
\definecolor{lightblue}{rgb}{0.68, 0.85, 0.9}

\usepackage{xspace}
\newcommand{\eg}{e.g.\xspace}
\newcommand{\ie}{i.e.\xspace}
\newcommand{\etal}{et al\xspace}
\begin{document}

\title{Training-Free Robust Interactive Video Object Segmentation}

\author{Xiaoli Wei, \thanks{Corresponding author: Chunxia Zhang. E-mail: cxzhang@mail.xjtu.edu.cn.} \thanks{Xiaoli Wei, Chunxia Zhang are with School of Mathematics and Statistics, Xi'an Jiaotong University, Xi'an, Shaanxi, 710049, China.}
\and
Zhaoqing Wang, \thanks{Zhaoqing Wang and Tongliang Liu are with the Sydney AI Center, School of Computer Science, Faculty of Engineering, The University of Sydney,
Darlington, NSW2008, Australia.}
\and
Yandong Guo, \thanks{Yandong Guo is with the School of Artificial Intelligence, Beijing University
of Posts and Telecommunications, Beijing 100876, China.}
\and
Chunxia Zhang, ~\IEEEmembership{Member, ~IEEE}, 
\and
Tongliang Liu, ~\IEEEmembership{Senior Member, ~IEEE},
\and
Mingming Gong, ~\IEEEmembership{Member, ~IEEE} \thanks{Mingming Gong is with the School of
Mathematics and Statistics, Faculty of Science, The University of Melbourne, Melbourne, VIC 3010, Australia.}
}

\markboth{Journal of \LaTeX\ Class Files,~Vol.~14, No.~8, August~2021}%
{Shell \MakeLowercase{\textit{et al.}}: A Sample Article Using IEEEtran.cls for IEEE Journals}


\maketitle

\begin{abstract}
    Interactive video object segmentation is a crucial video task, having various applications from video editing to data annotating.
    However, current approaches struggle to accurately segment objects across diverse domains.
    Recently, Segment Anything Model (SAM) introduces interactive visual prompts and demonstrates impressive performance across different domains.
    In this paper, we propose a training-free prompt tracking framework for interactive video object segmentation (I-PT), leveraging the powerful generalization of SAM.
    Although point tracking efficiently captures the pixel-wise information of objects in a video, points tend to be unstable when tracked over a long period, resulting in incorrect segmentation.
    Towards fast and robust interaction, we jointly adopt sparse points and boxes tracking, filtering out unstable points and capturing object-wise information.
    To better integrate reference information from multiple interactions, we introduce a cross-round space-time module (CRSTM), which adaptively aggregates mask features from previous rounds and frames, enhancing the segmentation stability.
    Our framework has demonstrated robust zero-shot video segmentation results on popular VOS datasets with interaction types, including DAVIS 2017, YouTube-VOS 2018, and MOSE 2023, maintaining a good tradeoff between performance and interaction time.
\end{abstract}


\section{Introduction}
\IEEEPARstart{V}{ideo} object segmentation (VOS) is a technique designed to segment dominant objects within video sequences. It plays a crucial role in advancing applications and developments in various fields, \eg, autonomous driving and robotics. According to the form of human intervention in the segmentation process, it can be mainly divided into three branches, including automatic video object segmentation (AVOS) \cite{yang2023multi}, semi-supervised video object segmentation (SVOS) \cite{cheng2022xmem}, and interactive video object segmentation (IVOS) \cite{cheng2021modular,zhou2022survey}. AVOS belongs to unsupervised segmentation methods, and humans do not participate in the inference process. SVOS usually requires manual annotation of the target object mask in the first frame. IVOS achieves manual-guided video segmentation to refine results and improve accuracy through multi-round interactions, establishing a human-in-the-loop VOS system.

With the advent of the public DAVIS interactive benchmark \cite{Perazzi2016,caelles20182018}, various approaches that focus on iteratively refining the segmentation workflow \cite{benard2017interactive,cheng2021modular} are introduced to enhance the efficiency of the data labeling process. Nevertheless, most of the prevailing methods \cite{heo2020interactive, miao2020memory, heo2021guided, cheng2021modular} adopt a non end-to-end framework that encompasses the interaction-to-mask process and mask propagation. This approach not only increases the complexity of the task but also exhibits limitations in processing novel data, particularly in scenarios requiring zero-shot segmentation across diverse situations.

Segment Anything Model (SAM) \cite{kirillov2023segment}, trained on the large-scale dataset, is a groundbreaking vision foundation model, demonstrating superior zero-shot segmentation capability. By resorting to that, recent methods \cite{yang2023track,cheng2023segment} have succeeded in generating initial segmentation masks with reduced reliance on manual annotations. However, their performance tends to falter across varied video domains, largely due to the limitations inherent in traditional video tracking models. SAM-PT \cite{rajivc2023segment} introduces point tracking into SAM, enhancing the zero-shot video segmentation across diverse video domains. This method propagates sparse points in videos to generate masks, avoiding dataset-specific fine-tuning and maintaining strong generalization. However, error accumulation from drift in point propagation can result in persistent segmentation inaccuracies, particularly in scenarios where objects occlude and reappear. This poses a substantial challenge in fully harnessing the potential of vision foundation models.

\begin{figure*}[t]
    \begin{center}
    \centerline{
        \includegraphics[width=\textwidth]{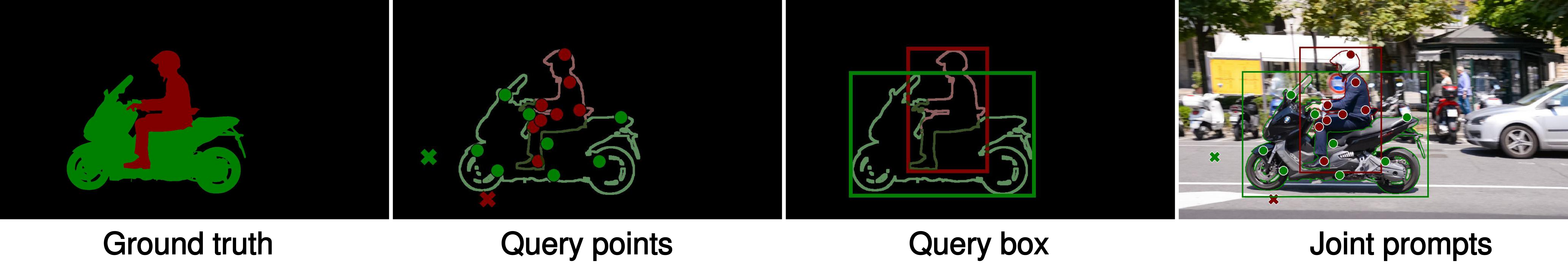}
    }

    \caption{Query point and box selection. Positive points are represented by circles, negative points by crosses, and different colors signify different objects. The edges of the target object are visualized to show where visual prompts are selected.}
    \label{fig:pis}
    \end{center}
\end{figure*}

To meet this challenge, in this paper, we propose an innovative training-free interactive video object segmentation framework, empowered by visual prompt tracking. As illustrated in Fig. \ref{fig:pis}, we utilize query boxes to indicate object locations, while query points are employed to capture the local structures of the segmented objects. Recognizing the susceptibility of points to drift during long-sequence tracking and the inadequacy of boxes for precise boundary delineation, our approach concurrently tracks query boxes and a user-defined set of query points on the unsatisfactory frames. Afterwards, we select confident points within the query boxes to accurately capture object-specific details. Given that SAM is tailored for image segmentation, we introduce the cross-round space-time module (CRSTM) to enhance spatio-temporal consistency. This module retains user interactions from multiple rounds and frames, adaptively propagating them to the rest of video frames within each interaction round. In doing so, we increasing the segmentation stability and achieve the better tradeoff between performance and efficiency.

With extensive experimental results, our proposed I-PT framework demonstrate strong capability in zero-shot IVOS with fewer interaction time, surpassing certain conventional IVOS models. Our framework is not sensitive to various prompt trackers. The training-free character retains SAM's generalization across diverse domains and can serve as a foundational IVOS framework. We summarise our contributions in the following aspects:
\begin{enumerate}
    \item We propose a training-free IVOS framework that integrate joint visual prompt tracking into SAM, remaining the strong generalization of vision foundation model.
    \item We associate the independent segmentation results across rounds and frames via CRSTM, enabling the spatio-temporal correspondence from reference frames to guide mask prediction.
    \item The synergy of the above mechanisms creates a generalize and efficient IVOS framework that demonstrates robust zero-shot segmentation capabilities across diverse domains.
\end{enumerate}

\section{Related Work}
\subsection{Video Object Segmentation}
Modern deep learning-based AVOS approaches aim to develop a universal video object representation, focusing on pixel instance embedding and integrating short-term and long-term information. Convolutional recurrent neural networks (RNNs) \cite{wang2019zero} and dual-stream optical flow techniques \cite {li2019motion, sun2018pwc} serve as paradigms for local temporal pattern delineation. Siamese-structured models \cite{lu2019see, lu2020zero} predominate in capturing long temporal correlations. Deep learning architectures for SVOS can be classified into propagation and matching-centric paradigms. Propagation-based methods \cite{perazzi2017learning, zhang2019fast, huang2020fast} utilize deep networks to indirectly capture motion details, facilitating the frame-by-frame transmission of segmentation from the initial annotation. The initial matching approach introduced, \ie, template match, leverages frames with annotations as models to explore alignment techniques \cite{chen2018blazingly, yang2021collaborative, voigtlaender2019feelvos}. Subsequently, methods utilizing attention mechanisms, representative space-time memory (STM) \cite{oh2019video} and space-time correspondence network (STCN) \cite{cheng2021rethinking}, have emerged as notably promising approaches. They use a memory module to archive past frame and mask information embedded in the memory networks, implementing a non-local attention mechanism for current mask prediction through memory matching. More improvements based on them are proposed, \eg, exploring more effective memory storage mechanisms \cite{liang2020video} and facilitating the collaborative association of multiple objects \cite{yang2021associating}.

The above feature matching relies on the memory network to obtain feature embedding. We obtain better powerful and generic features by leveraging the large vision model SAM.

\subsection{Interactive Video Object Segmentation}
Since Caelles \etal \cite{caelles20182018} introduced the round-based IVOS track, it has attracted many research works. They mainly aim to improve two main processes of multi-round interaction, including segmenting the current frame from simulated scribbles and temporal propagation of the annotation frame mask. The widely used methods can be classified into decoupled and coupled approaches. The first category \cite{benard2017interactive,oh2019fast, heo2020interactive} uses two separate networks. However, independent learning requires redundant feed-forward computations in each round, leading to inefficiency. The second category joins the two stages with interconnected encoders, and scribble features are propagated separately to the image encoding. STM is used in IVOS \cite{oh2020space} to realize the unified dissemination of spatio-temporal information. Miao \etal \cite{miao2020memory} integrate the interaction network and the communication network into a unified framework to directly avoid multiple rounds of feed-forward. Guided interactive segmentation (GIS) algorithm \cite{heo2021guided} addresses the interactive frame selection issue by introducing reliability-based attention maps, guiding users to quickly identify optimal frames. Yin \etal \cite{yin2021learning} argue that choosing the worst frame doesn't guarantee maximum progress and emphasis should be placed on the most valuable frame. Modular interactive VOS (MiVOS)\cite{cheng2021modular} incorporates difference-aware fusion into its decoupling framework, addressing the issue where the traditional linear weighting-based frame fusion process overlooks user intent.

Different from previous approaches, we propose a training-free framework, efficiently generalizing SAM from the image level to the video level.

\subsection{Vision Foundation Model}
Vision foundation models create a learning framework that interlinks data across various modalities \cite{awais2023foundational}. Enhanced by training on extensive datasets, it improves inference capability across multiple domains and exhibits powerful generalization, serving as a strong backbone for numerous downstream applications. The three mainstream branches are textual-prompt, visual-prompt, and generalist models. CLIP \cite{radford2021learning}, specialized for both image and text modalities and trained on datasets of a billion-scale, demonstrates top-leading potential in open-vocabulary image classification. SAM\cite{kirillov2023segment} integrates text and spatial prompts and establishes a generic feature extractor. HQ-SAM \cite{ke2024segment} enhances segmentation outcomes by incorporating high-quality tokens and training a new segmentation head while keeping the SAM backbone frozen. SegGPT \cite{wang2023seggpt} is a general model capable of performing arbitrary segmentation in both videos and images.

\begin{figure*}[!t]
	\centering
	\includegraphics[height=7.0cm]{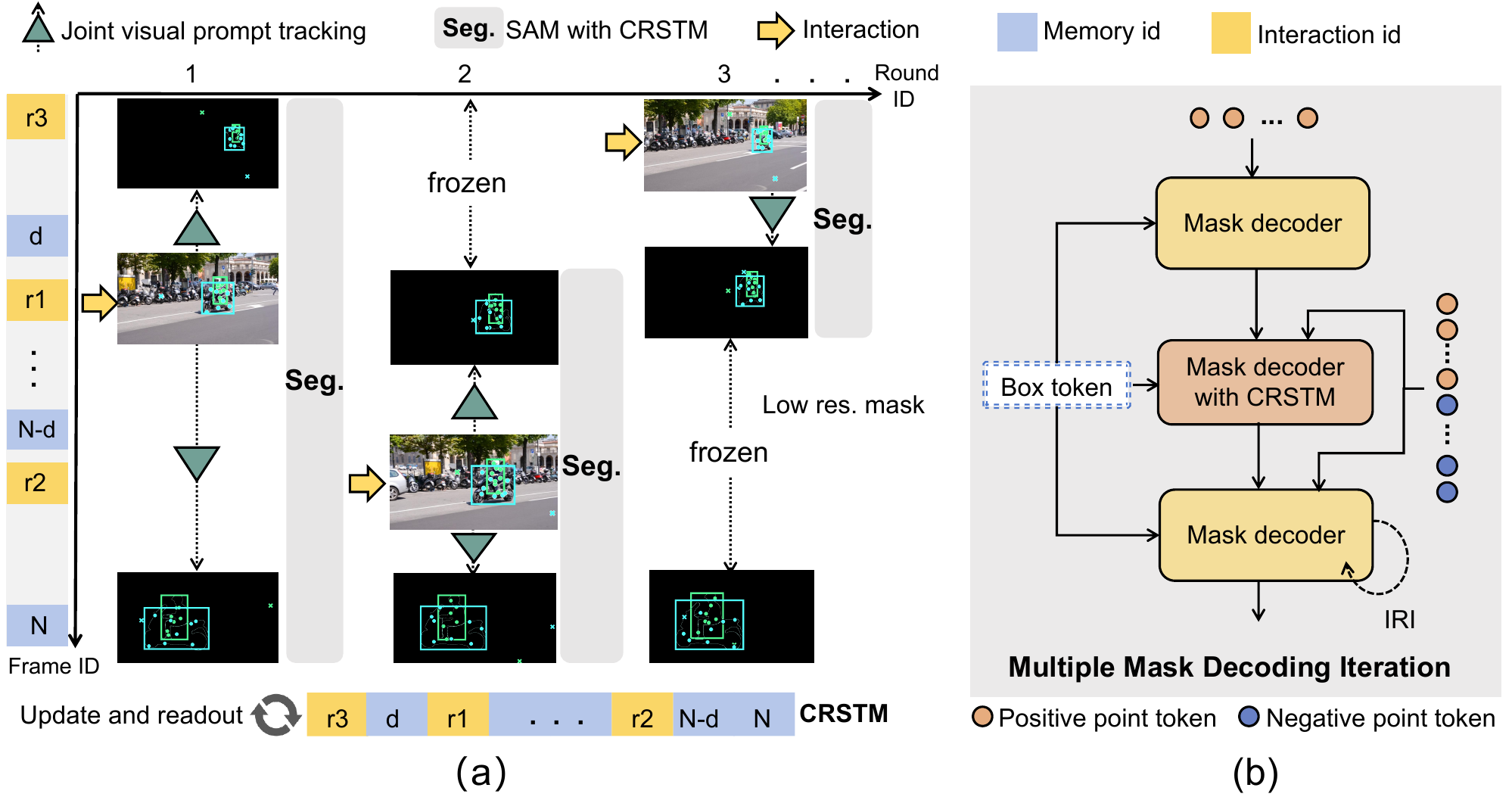}
	\caption{(a) Our proposed I-PT framework. Assuming the length of the video sequence is $N$. CRSTM stores information across frames at an interval of $d$. The interactive frames are denoted as $r$. (b) Multiple mask decoding iteration.
	}
	\label{fig:framework}

\end{figure*}
\section{Method}
First, Sec. \ref{sec:Overview of I-PT Framework} provides an overview of the proposed I-PT. Sec. \ref{sec:Prompt-Based Interactive System} presents the novel prompt-based interactive pipeline. Sec. \ref{sec:PT_and_IS} states the joint visual prompt tracking method and interactive segmentation. Finally, Sec. \ref{sec:CRSTM} describes CRSTM.

\subsection{Overview of I-PT’s Framework}
\label{sec:Overview of I-PT Framework}

Fig. \ref{fig:framework} shows the overall pipeline of our proposed method, which mainly consists of three components: a prompt-based interactive pipeline, joint visual prompt tracking, and segmentation. 
At the beginning, a prompt-based interactive pipeline selects a frame from a video sequence and gives prompts for target objects. The prompt tracker then propagates the prompts from the interactive frames to all video frames. Then, SAM with CRSTM predicts the target segmentation mask in the tracking range, and CRSTM updates the memory bank. The prediction for frames outside the range remains unchanged. The results for the whole video sequence are fed back to the prompt-based interactive pipeline, which assesses the segmentation quality across all video frames by comparing predictions to the ground truth, subsequently re-annotating the frames with the lowest accuracy. Later, the prompt tracker, SAM with CRSTM, and the prompt-based interactive pipeline form a feedback loop that iteratively refines the predicted mask. 

\subsection{Prompt-Based Interactive Pipeline}
\label{sec:Prompt-Based Interactive System}
The interactive mode reduces the cost of manual intervention compared to SVOS and provides higher flexibility than AVOS. While the scribble-based interactive track \cite{caelles20182018} lessens manual effort, it necessitates an additional step to train from scribble to mask. We modify the mode of interaction and build a more straightforward, simpler, and faster prompt-based interactive pipeline.

\subsubsection{Workflow.}
Our pipeline is based on a round-based process, allowing the first round of interaction frames not to be limited to the first frame of the video sequence, typically encompassing all objects to be segmented. Following the scribble-based interactive track \cite{caelles20182018}, we also used ground truth as a reference to facilitate simulation. Naturally, the actual operation lacks ground truth, relying instead on user selections.
First, the pipeline simulates user selections to provide prompts for objects in the interaction frame. After each segmentation process, the pipeline evaluates the entire video sequence by accessing the ground truth. It identifies the worst frame based on a user-defined optimization metric and then generates new prompts. 
Note that the interaction frames vary in each round.

\subsubsection{Points and Box Prompts.}
We simulate human annotation patterns to establish two types of interaction prompts, including query boxes and points, as illustrated in Fig. \ref{fig:pis}. We categorize query point attributes into positive points for target objects and negative points for non-target objects or backgrounds. 
We grid the target area for selecting positive points on a grid basis and randomly choose negative points from non-target areas. In this way, we avoid using any clustering functions and simply simulate human selection patterns. The query box is defined by the bounding rectangle around the target area, and we introduce an error within a specific pixel range to mimic human interaction. Finally, the joint prompts contain the query boxes and points.



\subsection{Joint Visual Prompt Tracking} 
\label{sec:PT_and_IS}
Point prompts capture pixel-wise structure, enabling better handling of complex contexts, yet they may become unstable and bring error accumulation over long periods, causing incorrect segmentation and intensive interaction.
Box prompts mark the region-wise structure, showing stability in long-distance propagation and helping filter noisy points. However, their coarse nature complicates distinguishing foreground and background in complex scenarios.
Thus, we integrate points and boxes for joint tracking.
This section focuses on prompt tracking, restricted propagation, and segmentation. We will introduce the details of CRSTM in the next section.

\subsubsection{Prompt Tracking.}
\label{sec:Prompt Tracking}
We use joint prompt tracking to propagate query points and boxes of the interaction frame to the video sequence, where points and boxes are fed into two different tracking pipelines. The point tracker PIPS \cite{harley2022particle} utilized in SAM-PT \cite{rajivc2023segment}, recognized for its better tracking capabilities, aligns precisely with our requirements, thus being incorporated into our point tracking pipeline to produce point trajectories and occlusion scores. Simultaneously, we observed strong robustness in our box tracking across various visual object tracking (VOT) methods. OSTrack \cite{ye2022joint} extracts discriminative, object-oriented features earlier within a highly parallelizable framework, setting it apart with rapid inference compared to strong competitors. After tracking, we remove points outside each frame's box to minimize noise impact. 
\subsubsection{Restricted Propagation.}
\label{sec:Restricted Propagation}
Temporal continuity in video sequences reveals a significant correlation between successive frames. This variation means that different interaction frames uniquely influence a query frame. Consequently, it is unnecessary to propagate prompts and update masks across the full timeline in each round. We thus choose a restricted propagation strategy \cite{oh2019fast}, where the bidirectional propagation of the current interaction frame ends once it reaches the closest previously interacted frame, respectively. As Fig. \ref{fig:framework} shows, except for the first round, updates in subsequent rounds are restricted to a specific tracking range, leaving the results in other frames frozen and unchanged. This configuration prevents target drift caused by distant interaction prompts and accelerates model inference. We denote the interaction frame of the $r$-th round as $t^r$, using superscripts to represent the round and subscripts to identify the frames. To update frames in each round, we utilize the following linear weighting formula
 \begin{equation}
 	M_i^r=M_i^{r-1}\frac{\left|t_i^r-t^r\right|}{\left|t^c-t^r\right|} + M_i^{r}\frac{\left|t_i^r-t^c\right|}{\left|t^c-t^r\right|},
 \label{eq:Restricted Propagation}
 \end{equation}
 where $t^r$ and $t^c$ represent the current interaction frame and the closest previous interacted frame in the $r$-th round. $t_i^r$ denotes the query frame in the $r$-th round. $M_i^r$ and $M_i^{r-1}$ signify the prediction masks for the individual $i$-th frame in the $r$-th round and $r-1$-th round, respectively.
\subsubsection{Segmentation.}
\label{sec:secSegmentation}
SAM \cite{kirillov2023segment} integrates three main parts, \ie, prompt
encoder, image encoder and lightweight mask decoder.
The prompt encoder converts different prompts into unique tokens. Round by round, it processes the outcomes of prompt tracking on a frame-by-frame basis, encoding the relevant prompt information. Image encoder is built on the MAE \cite{he2022masked} pre-trained Vision Transformer (ViT)\cite{dosovitskiy2020image}. Despite being a large and time-intensive model, it operates solely in the first round and requires only image embedding invocation in subsequent rounds. The mask decoder employs a transformer to map the prompt token features onto the image embedding. Since the complexity of point attributes, including occlusion, non-occlusion, positive, and negative types, we develop multiple mask decoding iterations, as depicted in Fig. \ref{fig:framework}. It diverges from the iterative refinement iterations (IRI) approach in \cite{rajivc2023segment}, which iteratively executes the mask decoder without incorporating the memory bank. The positive token exclusively embeds the positive non-occlusion points, while the remaining points are embedded as negative tokens.
Initially, the mask decoder processes only positive tokens and box tokens. On its second run, it accepts all prompt tokens and the first low-resolution mask, then outputs an enhanced low-resolution mask after CRSTM processing. Subsequently, it undergoes IRI to refine the mask.

\subsection{Cross-Round Space-Time Module (CRSTM)}
\label{sec:CRSTM}
Points and boxes are propagated throughout the sequence, yet SAM, being a 2D segmentation model, operates with predictions that are independent between frames in the segmentation stage. To further enhance the spatial-temporal correspondence across multiple interaction rounds and frames, we introduce CRSTM to leverage prior pair-wise reference information sufficiently.

\begin{figure*}[!t]
	\centering
	\includegraphics[height=7.0cm]{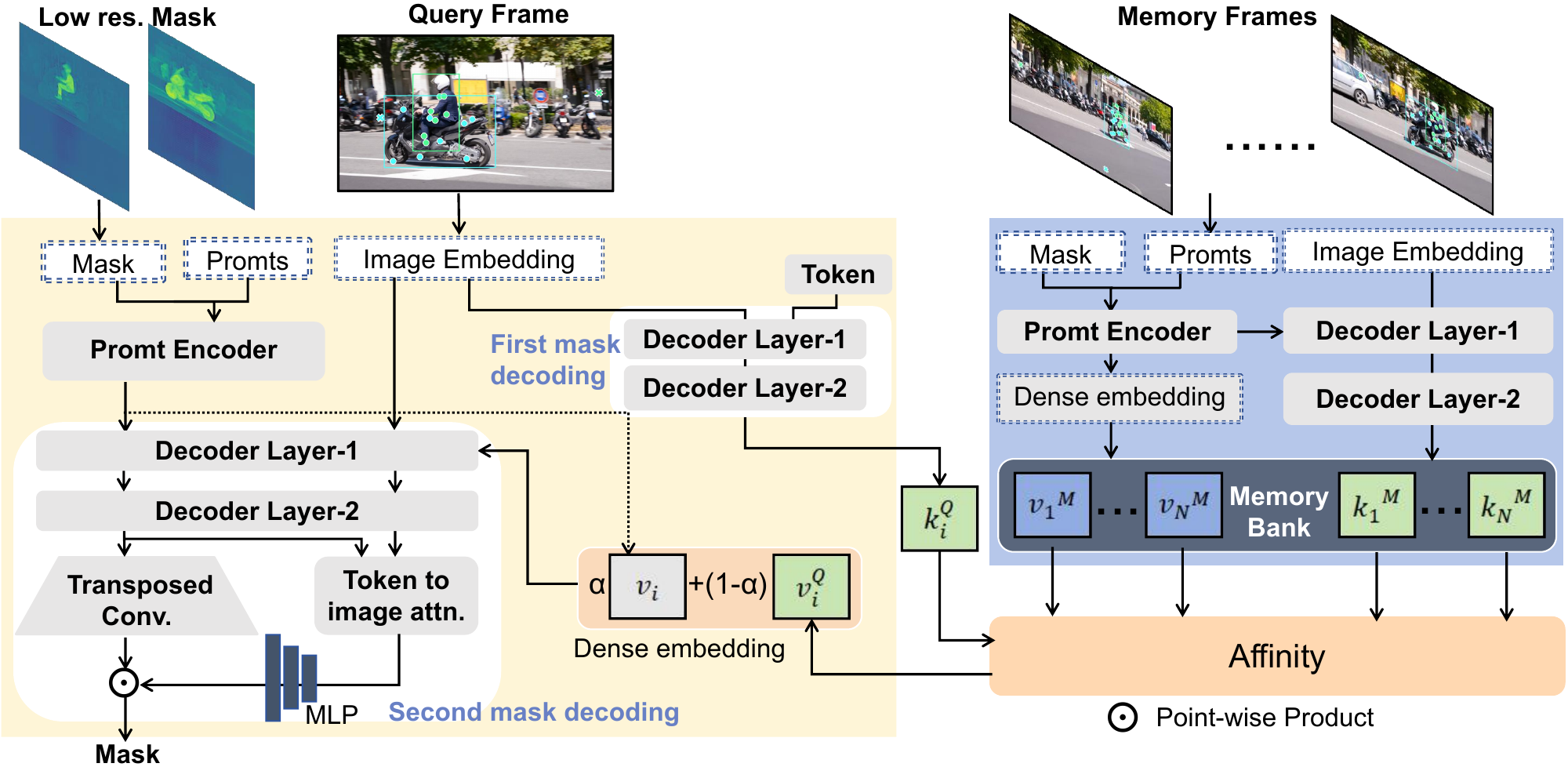}
	\caption{CRSTM architecture. It mainly contains three processes, \ie, memory updating, memory readout, and segmentation. }
	\label{fig:CRSTM}

\end{figure*}

Our memory bank employs a cross-frame storage strategy, with interaction frames from each round being sequentially recorded (Fig. \ref{fig:framework} shows an example in I-PT framework). As Fig. \ref{fig:CRSTM} illustrates, the mask decoder with CRSTM receives the query key from the first mask decoder output (referring to the multiple mask decoding iteration in Fig. \ref{fig:framework}). Next, it conducts time-space matching with the memory bank to retrieve the query value. Finally, this query value is combined with the original dense embedding through weighting, followed by the execution of this round's mask decoder. We denote the key and value as $\mathbf{k}$ and $\mathbf{v}$, respectively. The associated superscripts indicate whether they originate from a memory $M$ or query frame $Q$, and the subscripts represent the frame index. The index of the current query frame is $i$.

\subsubsection{Space-Time Correspondence.}
We observe that the dense embedding from SAM's prompt encoder encodes the mask feature. Meanwhile, the mask decoder executes token-to-image and image-to-token attention to update features, capturing both visual semantics and mask information. This renders them naturally fitting as keys and values. In selecting keys, we explore different strategies and discover that the feature output by the two-layer transformer module (Fig. \ref{fig:CRSTM}) in the mask decoder offers the most stability.

\subsubsection{Memory Updating.}
Before each segmentation round, the memory key and value within this round's propagation range are cleared, except for those of the interaction frame. Then, the key and value from the current round's interaction frames are stored in the memory bank before processing the other frames. Finally, our model executes segmentation from the interaction frame in both forward and backward directions. Upon segmenting each frame, the corresponding keys and values are updated if they fall within the memory index.

\subsubsection{Memory Readout.}
We can denote the current query frame with index $i$ and memory bank $ \{\mathbf{k}_1^M, \cdots, \mathbf{k}_N^M, \mathbf{v}_1^M, \cdots, \mathbf{v}_N^M\}$, where $\mathbf{k}_{.}^M, \mathbf{v}_{.}^M \in \mathbb{R}^{C \times H W}$, $C$ is the channel dimension (256 in SAM), and $H$ and $W$ are spatial dimensions (64 in SAM). Then, we can obtain the total memory key and memory value
\begin{equation}
\begin{aligned}
	\mathbf{K}^{M} = [\mathbf{k}_1^M, \cdots, \mathbf{k}_N^M ] \in \mathbb{R}^{C \times N H W}, \\
 \quad \mathbf{V}^{M} = [\mathbf{v}_1^M, \cdots, \mathbf{v}_N^M ] \in \mathbb{R}^{C \times N H W},
 \end{aligned}
\label{equ:k}
\end{equation}
where $[\: \cdot \:]$ symbolizes the operation of concatenating features along the spatial dimension. Given the query key $\mathbf{k}_i^Q \in \mathbb{R}^{C \times H W}$, we can get the pairwise affinity matrix $\mathbf{S} \in \mathbb{R}^{ N H W\times H W}$ by using the negative squared Euclidean distance in \cite{cheng2021rethinking} as
\begin{equation}
	\mathbf{S}_{j l}=-\left\|\mathbf{K}_j^M-\mathbf{K}_l^Q\right\|_2^2,
\end{equation}
where $\mathbf{K}_j^M$ and $\mathbf{K}_l^Q$ represent their feature vectors at the $j$-th position and $l$-th position, respectively.
The affinity matrix can be softmax-normalized as
\begin{equation}
	\mathbf{W}_{j l}=\frac{\exp \left(\mathbf{S}_{j l}\right)}{\sum_{m \in \operatorname{Top}_l^k(\mathbf{S})}\left(\exp \left(\mathbf{S}_{m l}\right)\right)}, \text{if  } j \in \operatorname{Top}_l^k(\mathbf{S}),
\end{equation}
where we also utilize top-$k$ filtering to extract highly relevant features. Next, we can acquire the following query value by applying matrix multiplication
\begin{equation}
	\mathbf{v}_i^Q=\mathbf{V}^M \mathbf{W} \in \mathbb{R}^{C \times H W},
\end{equation}
which carries more temporal consistent information from the space-time matching.  Meanwhile, the original value $\mathbf{v}_i$ (i.e., dense embedding in SAM) is only decoded from the current query frame, thus it aligns more closely with the current query frame. Owing to I-PT's restricted propagation (Sec. \ref{sec:Restricted Propagation}), the tracking range narrows as the interaction progresses. The guiding power of spatiotemporal information may also diminish accordingly. Considering this issue, we perform exponential weighting on the before and after query values based on the round index $r$ to calculate the final updated value as
\begin{equation}
	\mathbf{v}_i=\alpha \mathbf{v}_i + (1-\alpha) \mathbf{v}_i^Q, \quad \alpha=\frac{1}{1+\exp\left(-(r-R/2) / 2\right)},
\end{equation}
where $R$ represents the total round number.

\section{Experiments}
\label{sec:Experiments}
We first present the IVOS dataset and evaluation metric used in our study. Afterwards, We compare our framework with existing state-of-the-art IVOS methods. Finally, we conduct an ablation study to verify each component in our framework.
\subsection{Datasets}
\label{sec:Datasets}
{\bf{DAVIS 2017.}} DAVIS 2017 \cite{pont20172017} is a popular benchmark for previous IVOS studies, featuring a publicly assessable scribble-based interactive segmentation track \cite{caelles20182018}.
It includes challenging scenarios like partial or complete occlusion and the object's reappearance. Thus, our prompt-based interactive pipeline is developed using its validation set, which includes 30 video sequences. Each sequence offers 3 different initialization prompts, totaling 90 video sequences. Each video sequence undergoes 8 rounds of interaction, with the model's inference time for each object in each interaction capped at 30 seconds.

\noindent{\bf{YouTube-VOS 2018.}} We adopted the video interactive dataset constructed in \cite{yin2021learning}, which sampled 50 sequences from the YouTube-VOS 2018 training set \cite{xu2018youtube}. We initialized 3 different prompts for each video sequence, resulting in a total of 150 video sequences. All other settings remain consistent with those in setting DAVIS 2017.

\noindent{\bf{MOSE 2023.}} MOSE 2023 \cite{ding2023mose} is a dataset recently introduced, notable for its complex scenes, such as object disappearance-reappearance,
the existence of small or less noticeable entities, prolonged occlusion, and dense surroundings with multiple targets moving. Since the MOSE 2023 validation set lacks complete ground truth, we selected 50 video sequences from its training set and provided 3 initial prompts for each, creating 150 sequence samples in total. All other settings are the same with DAVIS 2017.

\subsection{Evaluation Metric}
\label{sec:Evaluation Metric}
The evaluation of IVOS focuses on the trade-off between the interaction time and segmentation result quality \cite{caelles20182018}. The interaction time includes the duration for the track to mimic human prompts and the execution time for the model to perform a new inference round. 
We adopt $\mathcal{J}$ at 60s ($\mathcal{J}^{\dagger}$), $\mathcal{J} \& \mathcal{F}$ at 60s ($\mathcal{J} \& \mathcal{F}^{\dagger}$), the area under the curve (AUC) of the J score ($\text{AUC-}\mathcal{J}$), and that of the $\mathcal{J} \& \mathcal{F}$ score ($\text{AUC-}\mathcal{J} \& \mathcal{F}$). Frame evaluation in each round locates the worst frame based on $\mathcal{J} \& \mathcal{F}$ score. 

\subsection{Implementation Details}
\label{sec:Implementation Details}
Our I-PT models are implemented in Python with PyTorch. All evaluation experiments are conducted on a single NVIDIA A100 GPU.

\noindent{\bf{Pre-trained Model.}} In our experiments, we directly used open-source checkpoints from their respective methods, including four point trackers (PIPS \cite{harley2022particle}, RAFT \cite{teed2020raft}, TapNet \cite{doersch2022tap}), CoTracker \cite{karaev2023cotracker}, 
four visual object trackers (OSTrack \cite{ye2022joint}, ROMTrack \cite{cai2023robust}, SBT \cite{xie2022correlation}, SeqTrack \cite{chen2023seqtrack}), SAM \cite{kirillov2023segment}, and HQ-SAM)\cite{ke2024segment}.  Our all experiments utilized the ViT-H backbone of SAM. It's important to note that all pre-trained models are not trained on the video segmentation dataset and do not overlap with our evaluation data. Thus, all evaluations of our models are conducted in zero-shot scenarios.

\noindent{\bf{Model Setting.}}
We adopt 8 positive points and 1 negative point suggested in \cite{rajivc2023segment}. 
The tracking models for points and boxes ultimately utilized in our I-PT are PIPS and OSTrack, respectively. 

\begin{table}[!t]
    \renewcommand{\arraystretch}{1.0}
    \scriptsize
    \caption{Quantitative results in DAVIS 2017 validation set \cite{pont20172017} compared with the state-of-the-art IVOS methods. -: indicating that this metric is not collected. SAM-PT$^{*}$ is the interactive version of SAM-PT \cite{rajivc2023segment}.}

    \label{tab:ivos_d17}
    \centering
    \setlength{\tabcolsep}{5pt}
    \begin{tabular}{cccccc}
    \toprule
    & \multicolumn{4}{c}{ \bf DAVIS 2017 Validation} \\
    \bf Method& \bf Interaction & $\text{AUC-}\mathcal{J}$  &  $\mathcal{J}^{\dagger}$ & $\text{AUC-}\mathcal{J} \& \mathcal{F}$ & $\mathcal{J} \& \mathcal{F}^{\dagger}$ \\
    \midrule
    \multicolumn{5}{l}{ (a) trained on video segmentation data} \\
    \midrule
    IPN \cite{oh2019fast} & & 69.1 & 73.4 & 77.8 & 78.7 \\
    MANet \cite{miao2020memory} && 74.9 & 76.1 & 76.1 & 76.5 \\
    IVOS-W \cite{yin2021learning}&& - &- & 74.1 &-\\
    ATNet \cite{heo2020interactive} && 77.1 & 79.0 & 80.9 & 82.7\\
    STM  \cite{oh2020space} &Scribble& - & - & 83.9 & 84.8\\
    GIS \cite{heo2021guided} && 82.0 & 82.9 &85.6 & 86.6 \\
    MIVOS+STM \cite{cheng2021modular} && 84.9 & 85.4 & 87.9 & 88.5 \\
    MIVOS+STCN \cite{cheng2021modular}&&- &- &88.4 &88.8 \\
    \midrule
    \multicolumn{5}{l}{ (b) zero shot on video segmentation data} \\
    \midrule
    SAM-PT$^{*}$ \cite{rajivc2023segment} &Points & 78.5 & 80.8 & 80.9 & 83.2 \\
    \rowcolor{lightblue}
    I-PT(ours) &Joint prompts & 79.3 & 81.2 & 81.9 & 83.8 \\
    \bottomrule
\end{tabular}

\end{table}

\begin{table}[!t]
	\renewcommand{\arraystretch}{1.0}
	\scriptsize
	\caption{Quantitative results in YouTube-VOS 2018 \cite{xu2018youtube} and MOSE 2023 \cite{ding2023mose} dataset compared with the state-of-the-art IVOS methods. -: indicating that this metric is not collected. SAM-PT$^{*}$ is the interactive version of SAM-PT \cite{rajivc2023segment}. }

	\label{tab:ivos_y18}
	\centering
	\setlength{\tabcolsep}{8pt}
	\begin{tabular}{ccccc}
		\toprule
		\multirow{2}{*}{\bf Method} &  $\text{AUC-}\mathcal{J}$  &  $\mathcal{J}^{\dagger}$ & $\text{AUC-}\mathcal{J} \& \mathcal{F}$ & $\mathcal{J} \& \mathcal{F}^{\dagger}$ \\
		\cmidrule{2-5}
		& \multicolumn{4}{c}{ \bf YouTube-VOS 2018} \\
		\midrule
		\multicolumn{5}{l}{ (a) trained on video segmentation data} \\
		\midrule
		IVOS-W $+$ IPN \cite{yin2021learning} & - &- & 44.7 &-\\
		IVOS-W $+$ MANet \cite{yin2021learning} & - &- & 66.9 &-\\
		IVOS-W $+$ ATNet \cite{yin2021learning} & - &- & 75.4 &-\\
		\midrule
		\multicolumn{5}{l}{ (b) zero shot on video segmentation data} \\
		\midrule
		SAM-PT$^{*}$ \cite{rajivc2023segment} & 80.4 & 82.5 & 80.4 & 82.5 \\
            \rowcolor{lightblue}
		I-PT(ours) & 82.1 & 84.1 & 82.2 & 84.2 \\
		\midrule
		& \multicolumn{4}{c}{ \bf MOSE 2023} \\
		\midrule
		\multicolumn{5}{l}{ (b) zero shot on video segmentation data} \\
		\midrule
		SAM-PT$^{*}$ \cite{rajivc2023segment} & 67.5 & 63.6 & 70.1 & 66.5 \\
            \rowcolor{lightblue}
		I-PT(ours) & 69.6 & 63.6 & 72.5 & 66.8 \\
		\bottomrule
	\end{tabular}

\end{table}

\subsection{Comparison with State-of-the-art Methods}
\label{sec:Comparison with State-of-the-art Methods}
\begin{figure*}[t]
	\centering
    \subfloat[]{%
        \includegraphics[height=4.0cm]{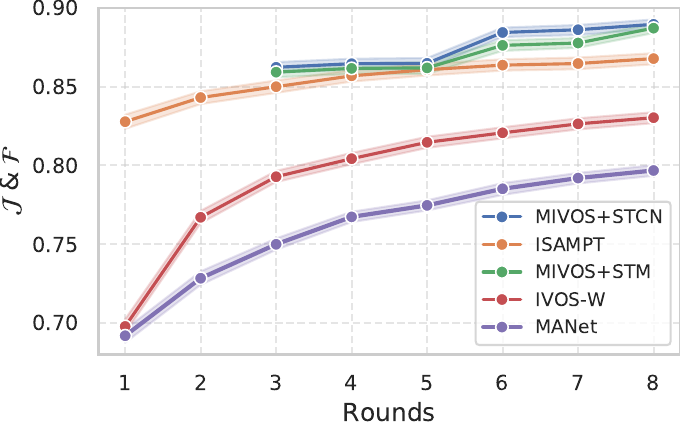}
        \label{fig:davis_ivos}
    }
    \hfill
    \subfloat[]{%
        \includegraphics[height=4.0cm]{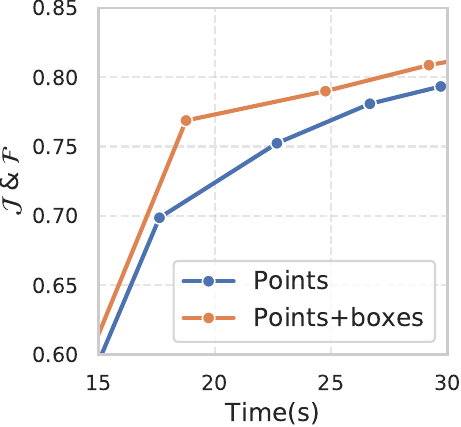}
        \label{fig:youtube_prompts}
    }
    \hfill
    \subfloat[]{%
        \includegraphics[height=4.0cm]{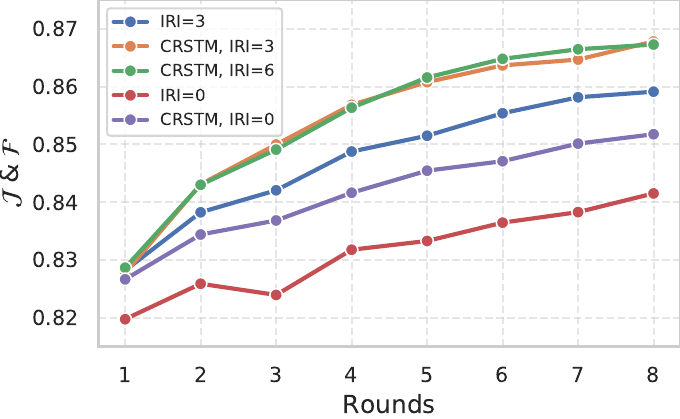}
        \label{fig:davis_j_and_f_crstm}
    }
    \caption{The curve of $\mathcal{J} \& \mathcal{F}$ versus interaction time and rounds. (a) Comparison of I-PT with some existing trained IVOS methods on DAVIS 2017 validation set \cite{pont20172017}. (b) Performance differences within different prompts on YouTube-VOS 2018 \cite{xu2018youtube}. (c) Performance differences across various configurations of the multiple mask decoding iteration on DAVIS 2017 validation set \cite{pont20172017}. }
	\label{fig:crstm_prompts}

\end{figure*}

\begin{figure*}[!t]
	\centering
    \subfloat[Results on the DAVIS 2017 validation set \cite{pont20172017}]{
        \includegraphics[height=4.01cm]{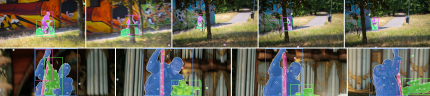}
        \label{fig:figure_DAVIS_2017_validation}
    }
    \hfill
    \subfloat[Results on the YouTube-VOS 2018 dataset \cite{xu2018youtube}]{
        \includegraphics[height=2.01cm]{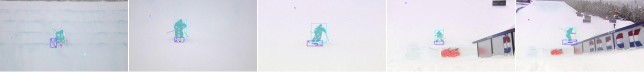}
        \label{fig:figure_YouTube-VOS_2018}
    }
    \hfill
    \subfloat[Results on the MOSE 2023 dataset \cite{ding2023mose}]{
        \includegraphics[height=5.22cm]{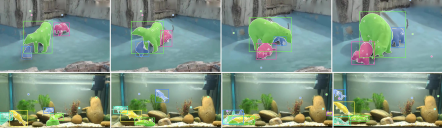}
        \label{fig:figure_MOSE_2023}
    }
	\caption{Visualization of I-PT object segmentation after 3 interactive rounds, with white circles for tracked positive points, crosses for negative points, and colored boxes for distinct objects.}
	\label{fig:figure results}

\end{figure*}

We collected results from existing state-of-the-art trained IVOS methods as reported in their respective publications. 
 Tab. \ref{tab:ivos_d17} and Tab. \ref{tab:ivos_y18} tabulate the quantitative comparisons across datasets DAVIS 2017, YouTube-VOS 2018, and MOSE 2023. First, it is obvious that our method significantly surpasses the baseline. Second, our I-PT method outperforms several existing trained IVOS methods, including IPN \cite{oh2019fast}, MANet \cite{miao2020memory}, IVOS-W \cite{yin2021learning}, and ATNet, marking a significant advancement in zero-shot IVOS. Fig. \ref{fig:davis_ivos} illustrates our method's progression in the $\mathcal{J} \& \mathcal{F}$ score across interaction rounds, showing I-PT's notable superiority over MANet \cite{miao2020memory} and IVOS-W \cite{yin2021learning} per round. Furthermore, I-PT can close the performance gap with MIVOS as interaction rounds increase. Fig. \ref{fig:figure results} visualizes the segmentation results of three rounds of interactive I-PT on three data sets, showing robust zero-shot segmentation capabilities. Our visualization frames avoid interaction frames. It can be found that our method can handle the segmentation of small objects and occlusion. In cases of high background and object similarity depicted in Fig. \ref{fig:figure_MOSE_2023}, our method yields segmentation capability yet faces challenges in clearly distinguishing between fish and stones in some frames. Additional visualization results are available in the supplementary material.

\subsection{Ablation Study}
\label{sec:Ablation Study}
We performed an ablation study on the DAVIS 2017 validation dataset, first examining various visual object trackers, point trackers, and segmentation models to confirm our method's robustness. Second, we study each component of the proposed I-PT model (adding  our proposed mechanisms in sequence) to assess its effectiveness. I-PT is a generic framework, and all pre-trained models can be selected to be more suitable and better based on field development and expectations.

\noindent{\bf{Robustness Evaluation.}}
Tab. \ref{tab:Robustness Evaluation} tabulates the results. Each visual object tracker features model variants tailored to specific input image pair resolutions. For instance, OSTrack-256 \cite{ye2022joint} operates with template and search regions measuring 128×128 pixels and 256×256 pixels, respectively, while another variant OSTrack-384 \cite{ye2022joint} uses 192×192 pixels for the template and 384×384 pixels for the search region. We evaluated each model's performance across various resolution variants, noting that despite minor differences among visual object trackers, all consistently outperformed the baseline (listed in Tab. \ref{tab:ivos_d17}) on the two AUC sores. HQ-SAM \cite{ke2024segment} does not perform as well as SAM, possibly due to it being a fine-tuned model that sacrifices some generalization capability. Overall, I-PT maintains stable predictive accuracy across the different configurations.

\begin{table}[!t]
	\renewcommand{\arraystretch}{1.0}
	\scriptsize
	\caption{Prompt tracker and segmentation model configuration ablation study results on the DAVIS 2017 validation set \cite{pont20172017}. PT: point tracker. BT: bounding box tracker. SEG: segmentation model. ${\dagger}$ Interpolated value @60s.
	}

	\label{tab:Robustness Evaluation}
	\centering
	\setlength{\tabcolsep}{1pt}
	\begin{tabular}{ccccccc}
		\toprule
		\multicolumn{3}{c}{\bf I-PT Configuration }  & \multicolumn{4}{c}{ \bf DAVIS 2017 Validation} \\
		PT & BT &SEG &$\text{AUC-}\mathcal{J}$  & $\mathcal{J}^{\dagger}$ & $\text{AUC-}\mathcal{J} \& \mathcal{F}$ & $\mathcal{J} \& \mathcal{F}^{\dagger}$ \\
		\midrule
		\multicolumn{6}{l}{ (a) point tracker } \\
		\midrule
		TapNet \cite{doersch2022tap} &  & & 76.1 & 77.3 & 78.4 & 79.8 \\
		RAFT \cite{teed2020raft} & OSTrack-256 \cite{ye2022joint}  & SAM \cite{kirillov2023segment}& 74.9 & 76.8 & 77.3 & 79.0 \\
		CoTracker \cite{karaev2023cotracker} &  & & 79.0 & 80.9 & 81.4 & 83.4 \\
		\midrule
		\multicolumn{6}{l}{ (b) bounding box tracker } \\
		\midrule
		& SeqTrack-b256 \cite{chen2023seqtrack} & & 78.8 & 80.1 & 81.3 & 82.5 \\
		& SeqTrack-b384 \cite{chen2023seqtrack} & & 78.8 & 80.4 & 81.3 & 82.9\\
		PIPS\cite{harley2022particle} &OSTrack-384 \cite{ye2022joint}& SAM \cite{kirillov2023segment}& 79.3 & 81.1 & 81.9 & 83.6 \\
		& ROMTrack-384 \cite{cai2023robust}& & 79.5 & 81.1 & 82.1 & 83.7\\
		& SBT-base \cite{xie2022correlation} & & 79.2 & 81.1 & 81.8 & 83.7 \\
		& ROMTrack-256 \cite{cai2023robust} & & 79.3 & 81.0 & 81.8 & 83.7 \\
		\midrule
		\multicolumn{6}{l}{ (c) segmentation model } \\
		\midrule
		PIPS \cite{harley2022particle}  & OSTrack-256 \cite{ye2022joint} &HQ-SAM \cite{ke2024segment} & 79.4 & 81.0 & 81.9 & 83.4 \\
		\midrule
            \rowcolor{lightblue}
		PIPS \cite{harley2022particle} & OSTrack-256 \cite{ye2022joint} &SAM \cite{kirillov2023segment} & 79.3 & 81.2 & 81.9 & 83.8 \\
		\bottomrule
	\end{tabular}

\end{table}

\noindent{\bf{{Framework Components.}}}
Tab. \ref{tab:DAVIS_ablation} illustrates the quantitative differences resulting from the stepwise evolutionary evolution of I-PT. We can find that the introduction of box tracking results in a 2.5-point improvement on $\mathcal{J} \& \mathcal{F}^{\dagger}$ due to its typically lower drift and higher stability. Fig. \ref{fig:youtube_prompts} shows a plot of $\mathcal{J} \& \mathcal{F}$ versus time for YouTube-VOS 2018. Adding a box prompt instead of merely adding points enables faster achievement of superior segmentation results. The addition of CRSTM further increased $\mathcal{J} \& \mathcal{F}^{\dagger}$ by 1.1 points. This indicates that historical spatio-temporal correspondence improves mask consistency over time. The iterative refinement continuously contributes a 0.6-point performance increase by leveraging information from the query frame itself to improve mask quality. Fig. \ref{fig:davis_j_and_f_crstm} displays $\mathcal{J} \& \mathcal{F}$ versus rounds on the DAVIS 2017 validation set. It's important to note that the total number of mask decoder iterations remains constant across experiments since CRSTM inherently includes one iteration (refering to Fig. \ref{fig:framework}). Observations reveal that, under an identical iteration count, the CRSTM strategy significantly enhances accuracy with increased interaction rounds. Increasing IRI appropriately leads to a notable boost in accuracy. As evidenced by Fig. \ref{fig:davis_j_and_f_crstm}, there's minimal difference between settings 3 and 6. Taking efficiency into account, as supported by Tab. \ref{tab:DAVIS_ablation}, setting IRI to 3 emerges as the optimal choice. The application of our CRSTM mechanism brings a better tradeoff between interaction time and segmentation accuracy. More results and discussions of ablation experiments can be found in the supplementary material. 

\begin{table}[t]
    \scriptsize
    \caption{I-PT configuration ablation study results on DAVIS 2017 validation set \cite{pont20172017}. ${\dagger}$ Interpolated value @60s. IRI: iterative refinement iterations. The score of gain is calculated from $\mathcal{J} \& \mathcal{F}^{\dagger}$.
	} 

	\label{tab:DAVIS_ablation}
	\centering
	\setlength{\tabcolsep}{5.0pt}
	\begin{tabular}{cccccccc}
		\toprule
		\multicolumn{3}{c}{\bf I-PT Configuration } &\multicolumn{5}{c}{\bf DAVIS 2017 validation set \cite{pont20172017}} \\
		\cmidrule{1-8}
		Prompt  &  CRSTM  & IRI  &  $\text{AUC-}\mathcal{J}$  &  $\mathcal{J}^{\dagger}$ & $\text{AUC-}\mathcal{J} \& \mathcal{F}$ & $\mathcal{J} \& \mathcal{F}^{\dagger}$ & Gain\\
		\midrule
		Point   & $\times$  & 0 & 75.7 & 77.9 & 77.5 & 79.6 &\\
		Point+Box & $\times$  & 0 & 77.8 & 79.6 & 80.2 & 82.1 & +2.5 \\
		\midrule
		Point+Box& $\checkmark$   &0 & 78.5 & 80.6 & 81.0 & 83.2 & +1.1 \\
		\midrule
		Point+Box  & $\checkmark$  & 1 & 78.8 & 81.1 & 81.3 & 83.7 &\\
            \rowcolor{lightblue}
		Point+Box  & $\checkmark$  & 3 & 79.3 & 81.2 & 81.9 & 83.8 & +0.6\\
		Point+Box  & $\checkmark$  & 6 & 79.3 & 81.0 & 82.0 & 83.6 & \\
		Point+Box  & $\checkmark$  & 12 & 79.2 & 80.6 & 81.8 & 83.3  &\\
		\bottomrule
	\end{tabular}
\end{table}

\section{Conclusion}
I-PT combines point tracking and visual object tracking to directly follow human-annotated points and boxes, leading to object segmentation through SAM. It employs cross-round space time feature matching, leveraging the temporal correlation between masks to enhance segmentation quality. Experimental results demonstrate I-PT's robust zero-shot generalizability in IVOS. I-PT seamlessly incorporates advanced tracking methods and large vision segmentation models without training, showcasing strong robustness and positioning it as a versatile foundational framework.

\bibliographystyle{IEEEtran}
\bibliography{main}







\end{document}